\documentclass{article}

\usepackage[whole]{bxcjkjatype}

\usepackage{PRIMEarxiv}

\usepackage[utf8]{inputenc} 
\usepackage[T1]{fontenc}    
\usepackage{hyperref}       
\usepackage{url}            
\usepackage{booktabs}       
\usepackage{amsfonts}       
\usepackage{nicefrac}       
\usepackage{microtype}      
\usepackage{lipsum}
\usepackage{fancyhdr}       
\usepackage[pdftex]{graphicx}       
\usepackage{subcaption}
\usepackage{authblk}
\usepackage{amsmath}
\usepackage{bm}
\usepackage{indentfirst}
\graphicspath{{media/}}     

\title{Techniques for Enhancing Memory Capacity of Reservoir Computing
}

\author[1]{Atsuki Yokota}
\author[1]{Ichiro Kawashima}
\author[1]{Yohei Saito}
\author[1,2]{Hakaru Tamukoh}
\author[1,2,3]{Osamu Nomura}
\author[1,2]{Takashi Morie}

\affil[1]{Graduate School of Life Science and Systems Engineering, Kyushu Institute of Technology, Japan}
\affil[2]{Research Center for Neuromorphic AI Hardware, Kyushu Institute of Technology, Japan}
\affil[3]{Faculty of Informatics, The University of Fukuchiyama, Japan}

\begin{document}
\maketitle

\vspace{-13mm}

\begin{abstract}
Reservoir Computing (RC) is a bio-inspired machine learning framework, and various models have been proposed.
RC is a well-suited model for time series data processing, but there is a trade-off between memory capacity and nonlinearity.
In this study, we propose methods to improve the memory capacity of reservoir models by modifying their network configuration except for the inside of reservoirs.
The \textit{Delay} method retains past inputs by adding delay node chains to the input layer with the specified number of delay steps.
To suppress the effect of input value increase due to the \textit{Delay} method, we divide the input weights by the number of added delay steps.
The \textit{Pass through} method feeds input values directly to the output layer. The \textit{Clustering} method divides the input and reservoir nodes into multiple parts and integrates them at the output layer.
We applied these methods to an echo state network (ESN), a typical RC model, and the chaotic Boltzmann machine (CBM)-RC, which can be efficiently implemented in integrated circuits.
We evaluated their performance on the NARMA task, and measured information processing capacity (IPC) to evaluate the trade-off between memory capacity and nonlinearity.
\end{abstract}


\section{Introduction}
In recent years, the rapid advancement of artificial intelligence (AI) technologies has garnered significant attention in various fields and has profoundly impacted society.
Among these technologies, reservoir computing (RC) has attracted considerable interest in time series data processing due to its high computational performance and simple learning algorithms.

Two critical properties of RC are its memory capacity, which indicates how long past inputs can be retained, and its nonlinearity, which describes the ability to map input time series into output sequences nonlinearly. 
It is well-known that these two properties generally exhibit a trade-off relationship \cite{Inubushi2017}. 
Since memory capacity and nonlinearity vary depending on the RC model, it is crucial to select an appropriate model based on the requirements of the target task.

This study aims to enhance the memory capacity of RC by modifying the network configuration without changing the RC model itself.
We propose novel methods to achieve this objective, ensuring that the RC model has sufficient memory capacity for the target task. 
The effectiveness of the proposed methods in improving memory capacity is evaluated using the NARMA task. 
We also investigate how the trade-off between memory capacity and nonlinearity in RC changes by analyzing the information processing capacity (IPC) \cite{Dambre2012}.

\section{Proposed methods}
\label{sec:methods}
In this paper, we propose three methods to enhance the memory capacity of reservoir computing (RC). 
The memory capacity ${\rm MC}$ is an index that quantifies how long an RC can store the input sequence, and is defined by 

\begin{equation}
 {\rm MC} = \sum_{T}Cor(T)^2=\sum_{T}\frac{Cov[y_{\mathrm{target}}(n,T), y_{\mathrm{out}}(n)]^2}{Var[y_{\mathrm{target}}(n,T)]Var[y_{\mathrm{out}}(n)]}
\end{equation}

\begin{figure}[ht]
    \centering
    \begin{subfigure}{0.32\textwidth}
        \centering
        \includegraphics[width=\linewidth]{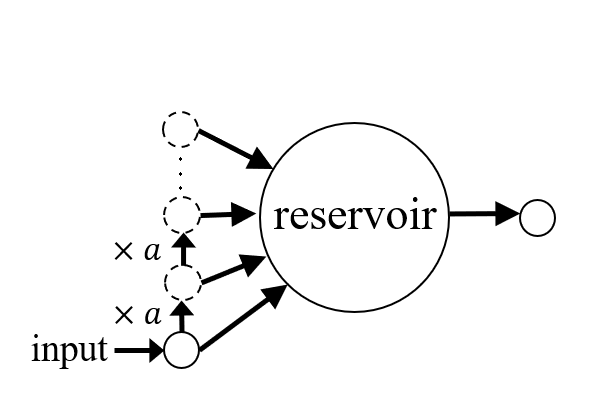}
        \caption{\textit{Delay}}
        \label{fig:delay}
    \end{subfigure}
    \begin{subfigure}{0.32\textwidth}
        \centering
        \includegraphics[width=\linewidth]{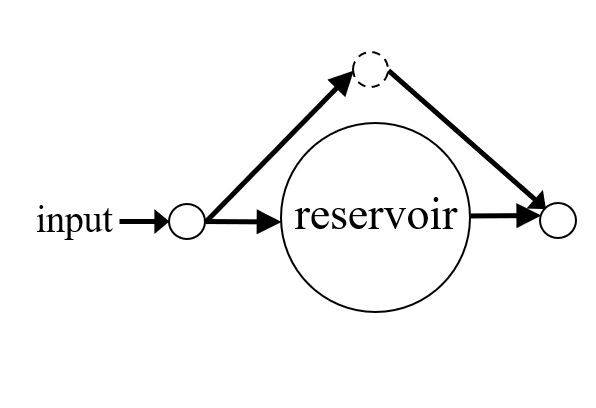}
        \caption{\textit{Pass through}}
        \label{fig:passthrough}
    \end{subfigure}
    \begin{subfigure}{0.32\textwidth}
        \centering
        \includegraphics[width=\linewidth]{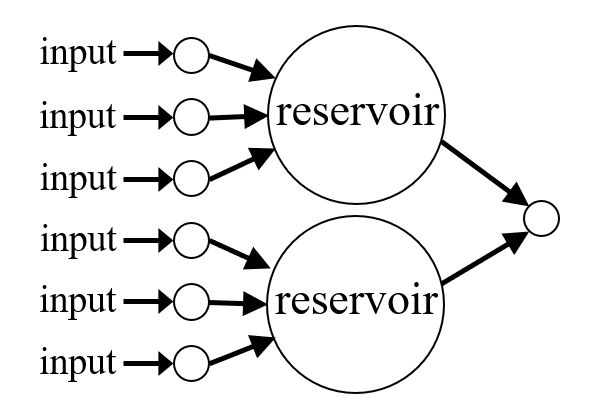}
        \caption{\textit{Clustering}}
        \label{fig:clustering}
    \end{subfigure}
    \caption{Proposed methods.}
    \label{fig:proposed_methods}
\end{figure}

\begin{figure}[ht]
    \centering
    \begin{subfigure}{0.32\textwidth}
        \centering
        \includegraphics[width=\linewidth]{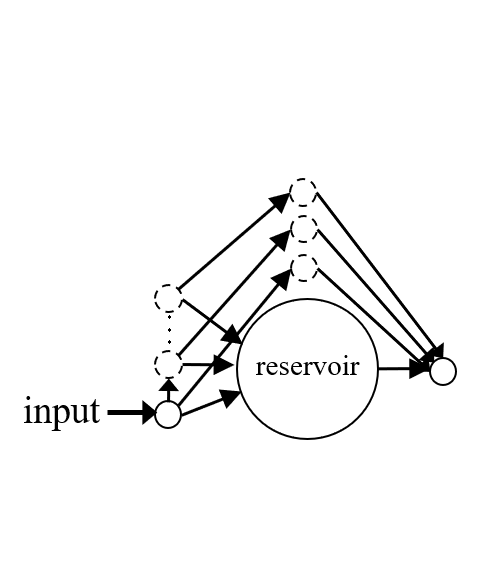}
        \caption{\textit{Delay}-\textit{Pass through}}
        \label{fig:delay}
    \end{subfigure}
    \begin{subfigure}{0.32\textwidth}
        \centering
        \includegraphics[width=\linewidth]{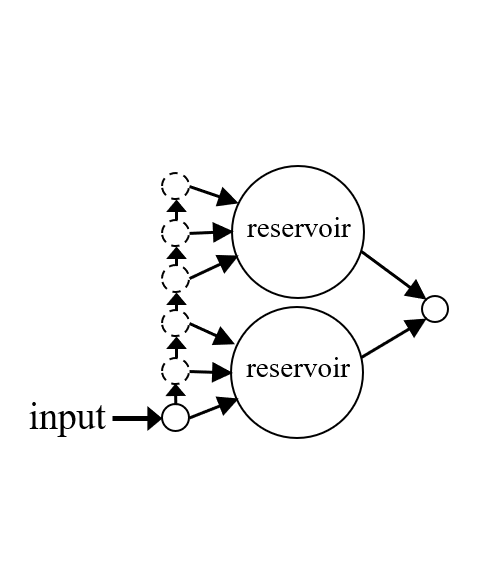}
        \caption{\textit{Delay}-\textit{Clustering}}
        \label{fig:passthrough}
    \end{subfigure}
    \begin{subfigure}{0.32\textwidth}
        \centering
        \includegraphics[width=\linewidth]{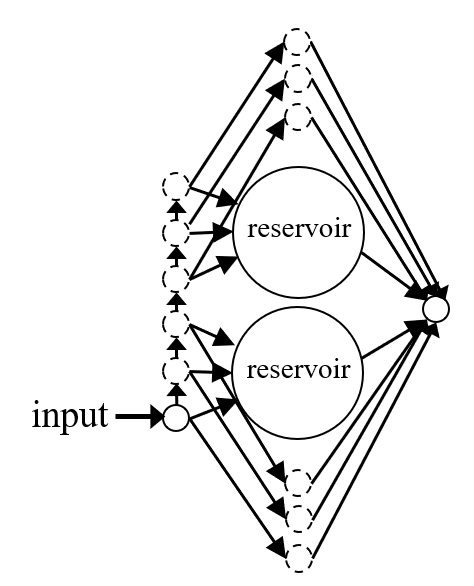}
        \caption{\textit{Delay}-\textit{Pass through-\textit{Clustering}}}
        \label{fig:clustering}
    \end{subfigure}
    \caption{Examples of combining proposed methods.}
    \label{fig:methods_combination}
\end{figure}

\noindent
Here, $T$ represents the number of input delay steps, $y_{\mathrm{target}}(n,T)$ denotes the target signal, and $y_{\mathrm{out}}(n)$ represents the output signal. 
The $Cor$, $Cov$, and $Var$ denote correlation, covariance, and variance, respectively. Additionally, $Cor(T)^2$ is also referred to as the coefficient of determination.
The proposed methods are illustrated in Figure \ref{fig:proposed_methods}.
In addition, the proposed methods can also be combined, as shown in Figure \ref{fig:methods_combination}.

\subsection{\textit{Delay} method}
The \textit{Delay} method retains past inputs by adding delay node chains to the input layer with the specified number of delay steps and by propagating past inputs between the nodes of the input layer.
This method duplicates the input layer nodes for the desired number of time steps. It propagates the past input across the duplicated input layer nodes, multiplying by a decay factor $a \ (0 < a \leq 1)$ to retain the past input information within the input layer.
Since the number of reservoir input nodes increases due to the \textit{Delay} method, the influence of the input on the reservoir increases, potentially causing the values of activation functions of the reservoir nodes to reach near-saturation regions and degrading the reservoir performance. 
To mitigate this, a scaling factor $s$ is applied to the input weight $\bm{W}_\mathrm{in}$, which depends on the input dimension $N_\mathrm{in}$ and the number of delay steps $d$ based on the \textit{Delay} method, suppressing the increase in input magnitude; $s$ is given below.

\begin{equation}
s = \frac{1}{N_\mathrm{in} d}
\end{equation}

\subsection{\textit{Pass through} method}
The \textit{Pass through} method adds intermediate layer nodes that correspond one-to-one with the input layer nodes and directly provide input values to the output layer. By bypassing the nonlinear transformation of the reservoir layer, this method is expected to be particularly effective for tasks with high linearity.

\subsection{\textit{Clustering} method}
The \textit{Clustering} method partitions the input layer and reservoir layer nodes into multiple clusters and integrates them in the output layer. This method allows each cluster to acquire different dynamics. Notably, when combined with the \textit{Delay} method, it is expected that the clusters can capture various dynamics, ranging from those formed by distant past inputs to those formed by recent inputs.

\section{Reservoir models for evaluation}
\label{sec:model}
The proposed methods were evaluated by numerical simulation using two typical RC models: the echo state network (ESN) \cite{Jaeger2004}, 
which has a large memory capacity and low nonlinearity, and the chaotic Boltzmann machine (CBM)-RC \cite{Suzuki2013,Katori2019}, 
which exhibits a small memory capacity and high nonlinearity.

\subsection{ESN}
Let the $i$-th $(i = 1, 2,\ldots,N_\mathrm{rec})$ internal state of the ESN neuron used in this study be denoted as $x^{\mathrm{ESN}}_i(n)$ and the $j$-th $(j = 1, 2,\ldots,N_\mathrm{in})$ input from the input layer be denoted as $u_j(n)$. 
The update equation of the ESN is given by Eq.\ref{eq:esn_x}.

\begin{equation}
\label{eq:esn_x}
x^{\mathrm{ESN}}_i(n+1) = f(\sum_{j=1}^{N_\mathrm{in}}w^{\mathrm{in}}_{ij}u_j(n)+\sum_{j=1}^{N_\mathrm{rec}}w^{\mathrm{rec}}_{ij}x^{\mathrm{ESN}}_j(n))
\end{equation}

\noindent
Here, $\bm{W}_\mathrm{in}=[w^{\mathrm{in}}_{ij}]$ is initialized with uniform random values in the range $\left[-1, 1\right]$, and then scaled by the input intensity parameter $\alpha_\mathrm{in}$. 
The reservoir weight matrix $\bm{W}_\mathrm{rec}=[w^{\mathrm{rec}}_{ij}]$ is constructed by randomly placing three values $\left\{-1, 0, 1 \right\}$ within the matrix with a density  $\beta_\mathrm{rec}$. After initialization, multiplied by a constant so that the largest eigenvalue of the matrix is 1, followed by scaling with the spectral radius parameter $\alpha_\mathrm{rec}$. 
The function $f$ is the activation function, which is typically chosen as the hyperbolic tangent function.

\subsection{CBM-RC}
The discrete-time input and output variables are encoded into and decoded from the continuous-time pulse signals of CBM-RC, respectively.
Let the $i$-th $(i = 1, 2,\ldots,N_\mathrm{rec})$ internal state of CBM neurons used in this study be denoted as $x^{\mathrm{CBM}}_i(t)$, the binary output as $S_i(t)$, the input to the neuron as $Z_i(t)$, and the temperature as $T_c$. The update equations of CBM-RC are given below.

\begin{equation}
\label{eq:cbm_dxdt}
\frac{dx^{\mathrm{CBM}}_i(t)}{dt} = (1-2S_i(t))(1-\exp{\frac{(1-2S_i(t))(Z_i(t)J_i(t))}{T_c}})
\end{equation}

\begin{equation}
\label{eq:cbm_zi}
Z_i(t)=\sum_{j=1}^{N_\mathrm{in}}w^{\mathrm{in}}_{ij}(2u^{(s)}_j(t)-1)+\sum_{j=1}^{N_\mathrm{rec}}w^{\mathrm{rec}}_{ij}(2S_j(t)-1)
\end{equation}

\begin{equation}
\label{eq:cbm_si}
S_i(t) =
  \begin{cases}
    0 & \mathrm{when} \ (x^{\mathrm{CBM}}_i(t)=0) \\
    1 & \mathrm{when} \ (x^{\mathrm{CBM}}_i(t)=1) \\
  \end{cases}
\end{equation}

\begin{equation}
\label{eq:cbm_ji}
J_i(t)=\alpha_i(S_i(t)-S_{\mathrm{ref}}(t))(2S_i(\lfloor t \rfloor -1)
\end{equation}

\vspace{2mm}

\noindent
The weight matrices $\bm{W}_\mathrm{in}$ and $\bm{W}_\mathrm{rec}$ were initialized following the same procedure as in the ESN.
Here, $u^{(s)}_i(t)$ is the binary sequence $u^{(s)}_i(t) \in \{0,1\}$ encoded from the input layer and is given by Eq. \ref{eq:cbm_us}.
The continuous value of $u_i(t)$ is represented by the phase difference between $u^{(s)}_i(t)$ and $\Phi(t)$.

\begin{equation}
\label{eq:cbm_us}
u^{(s)}_i(t)=\Phi(t-\frac{1}{2}u_i(\lfloor t \rfloor))
\end{equation}

\noindent
where $\Phi(t)$ is pulse wave function: $\Phi(t)=\Theta(\sin(2\pi t))$, and $\Theta(x)$ is the Heaviside step function.
In CBM-RC, internal dynamics is controlled by the reference clock signal $S_\mathrm{ref}(t)=\Phi(t)$, ensuring the echo state property required as a reservoir function. 
The parameter $\alpha_i$ is introduced to adjust the intensity of $S_\mathrm{ref}(t)$.

\section{Numerical simulation and results}
\label{sec:exp}
In the subsequent numerical simulations, we set $N_\mathrm{in} = 1$, $N_\mathrm{rec} = 200$, and $N_\mathrm{out} = 1$.

\subsection{Evaluation Using NARMA task}
The effectiveness of the proposed methods was validated using the non-linear autoregressive moving average (NARMA) task as a benchmark. The NARMA task is widely used as an RC benchmark that models the relationship between the input and output time-series data using the following nonlinear difference equation, which incorporates time delays:

\begin{equation}
y(n+1) = \alpha y(n)+ \beta y(n) ( \sum^{T}_{m=0} y(n-m) )+ \gamma u(n-T+1)u(n) + \delta
\end{equation}

\noindent
where the input distribution is defined as $u(t) \sim \rm{Uniform}(0.0, 0.5)$, and the parameters are set to $(\alpha, \beta, \gamma, \delta) = (0.3, 0.05, 1.5, 0.1)$. The number of input delay steps $T$ was varied from 0 to 15, and the coefficient of determination was measured for each case.
The parameters for each RC were optimized using Bayesian optimization as shown in Table \ref{table:narma_parameters}.

\begin{figure}[t]
    \centering
    \begin{subfigure}{0.49\textwidth}
        \centering
        \includegraphics[width=\linewidth]{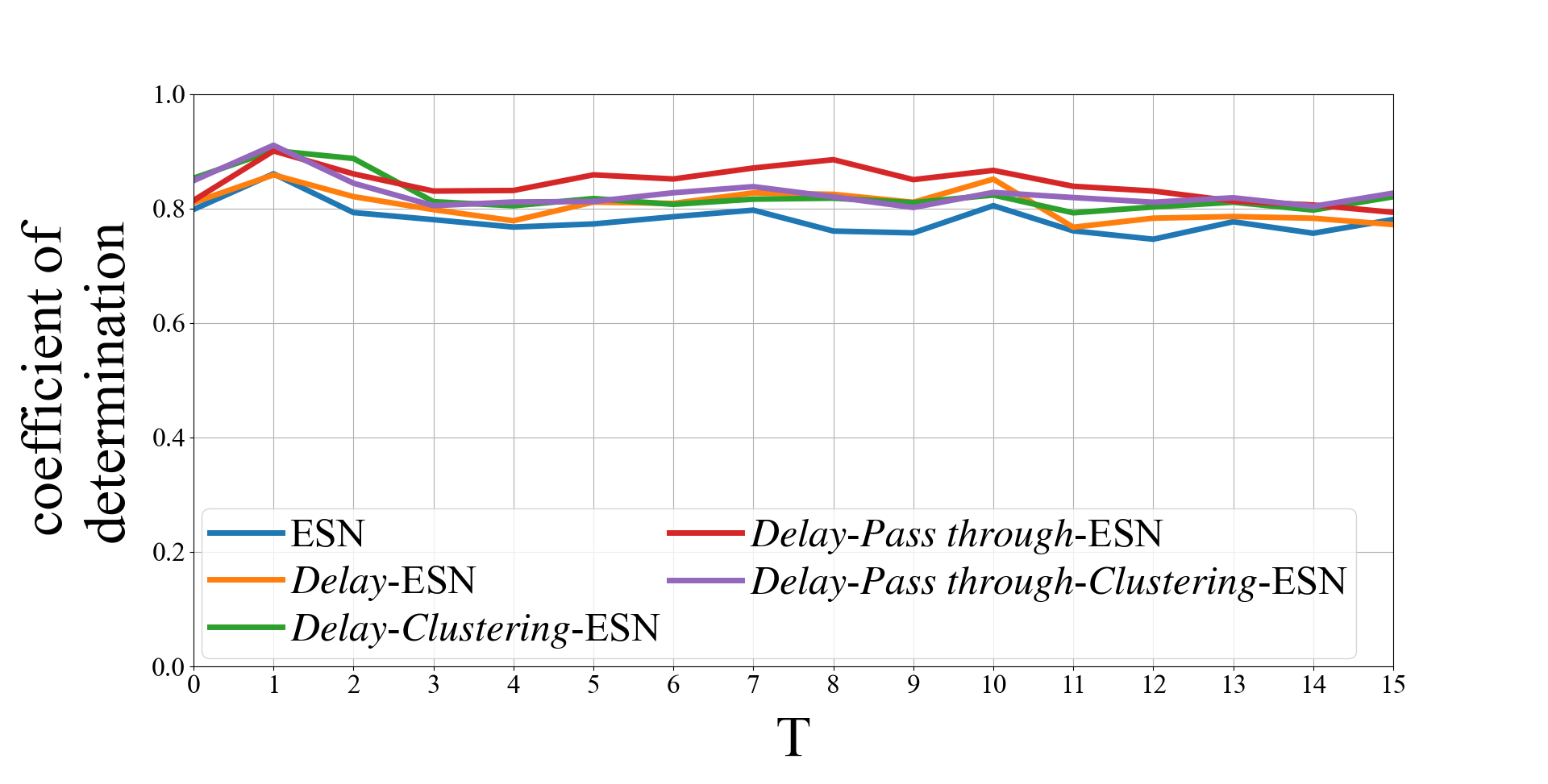}
        \caption{ESN}
        \label{fig:narma_esn}
    \end{subfigure}
    \begin{subfigure}{0.49\textwidth}
        \centering
        \includegraphics[width=\linewidth]{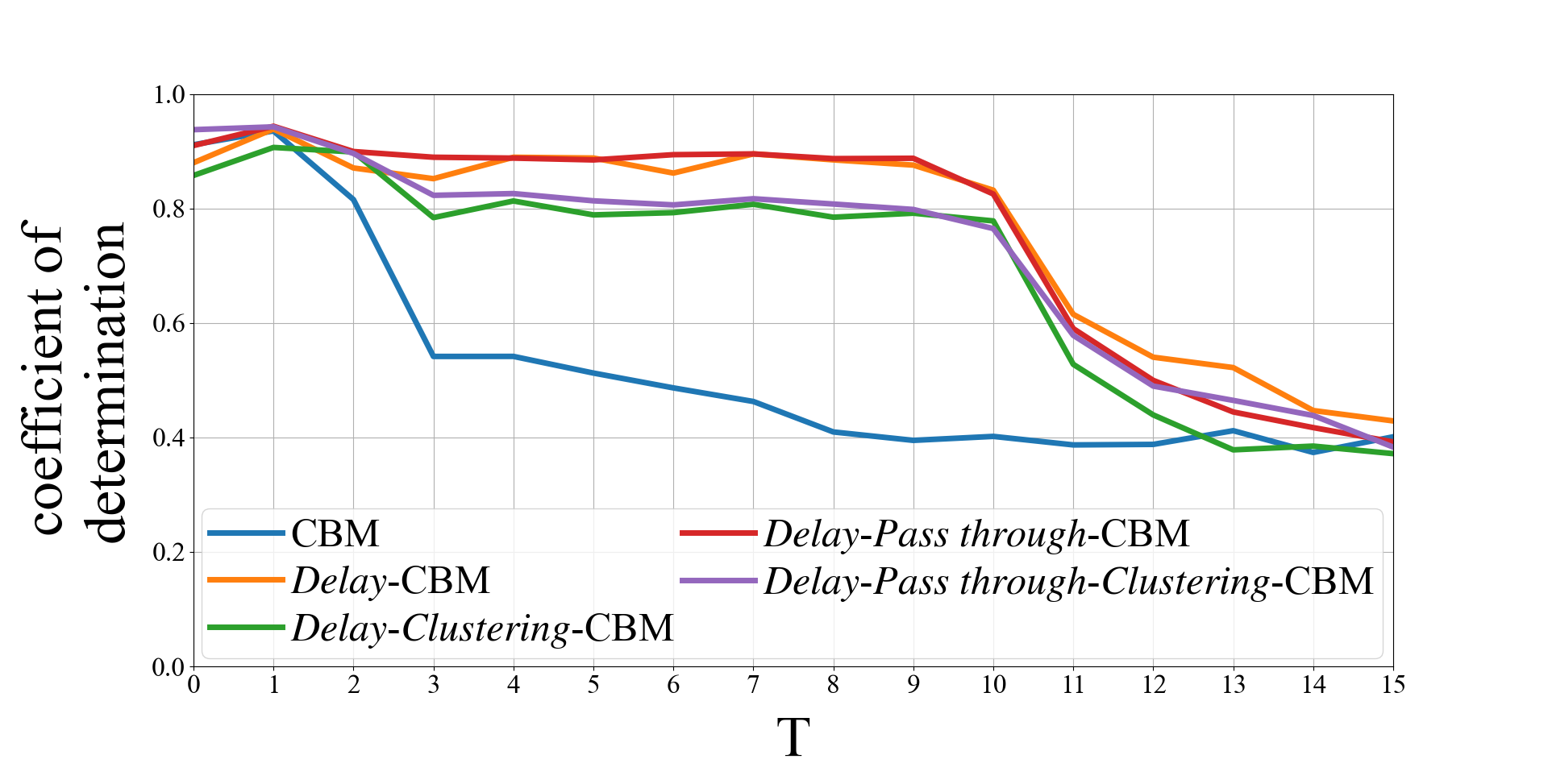}
        \caption{CBM-RC}
        \label{fig:narma_cbm}
    \end{subfigure}
    \caption{Evaluation results for NARMA tasks.}
    \label{fig:narma_results}
\end{figure}

The evaluation results are shown in Figure \ref{fig:narma_results}. 
The integral of the coefficient of determination graph represents the memory capacity. 
It can be observed that the proposed methods contribute to enhancing the memory capacity for both ESN and CBM-RC models compared to their original RC counterparts, with \textit{Delay} method making the most significant contribution. 
When solving relatively linear tasks, such as the NARMA, the performance of the CBM-RC model, which has a high degree of nonlinearity, is improved due to \textit{Delay} and \textit{Pass through} methods, which enhance the linearity of the model.

\begin{table}[t]
\centering
\caption{Parameters for evaluation with NARMA tasks}
\label{table:narma_parameters}
\begin{tabular}{llllllll}
\hline
RC model                            & $\alpha_\mathrm{in}$ & $\alpha_\mathrm{rec}$ & $\beta_\mathrm{rec}$ & $\alpha_i$ & $d$ & $a$ & $m$  \\ \hline \hline
ESN                                 & 0.9125               & 1.104                 & 0.3139               & -  & 10 & 1.0 & 5         \\
\textit{Delay}-ESN & 0.8668               & 0.8261                & 0.2126               & -  & 10 & 1.0 & 5         \\
\textit{Delay}-\textit{Pass through}-ESN & 0.9633               & 0.8310                & 0.2664               & -  & 10 & 1.0 & 5         \\
\textit{Delay}-\textit{Clustering}-ESN & 0.6534               & 1.102                 & 0.1638               & - & 10 & 1.0 & 5          \\
\textit{Delay}-\textit{Pass through}-\textit{Clustering}-ESN & 0.4871               & 1.110                 & 0.2423               & - & 10 & 1.0 & 5          \\
CBM-RC                              & 0.4321               & 0.5476                & 0.7687               & 0.5954 & 10 & 1.0 & 5    \\
\textit{Delay}-CBM & 0.2371               & 0.1641                & 0.5979               & 0.3718 & 10 & 1.0 & 5    \\
\textit{Delay}-\textit{Pass through}-CBM & 0.5618               & 0.1833                & 0.6356               & 0.4254 & 10 & 1.0 & 5    \\
\textit{Delay}-\textit{Clustering}-CBM & 0.2589               & 0.1221                & 0.0521               & 0.4339 & 10 & 1.0 & 5    \\
\textit{Delay}-\textit{Pass through}-\textit{Clustering}-CBM & 0.5379               & 0.4293                & 0.6495               & 0.3664 & 10 & 1.0 & 5 \\ \hline  
\end{tabular}
\end{table}

\subsection{Evaluation Using IPC}
To further analyze the effects of the proposed methods, particularly the \textit{Delay} method, which significantly contributed to enhancing the memory capacity in the NARMA task evaluation, we measured the information processing capacity (IPC), which is defined by the mean squared error normalized by the target output, 
\begin{equation}
 {\rm IPC} 
 = \lim_{N \to \infty} 
 \frac{\min_W \frac{1}{N} \sum_{n=1}^N \| y(n) - W x(n) \|^2 }{ \frac{1}{N} \sum_{n=1}^N \| y(n) \|^2}. 
\end{equation}
This assesses how the proposed method affected both the nonlinearity and the memory capacity of RC.
IPC was estimated using asymptotic expansion and the least squares method, following the approach in \cite{Saito2025}.
The RC models were trained using the data generated under the conditions shown in Table \ref{table:ipc_parameters}, and the IPC was estimated based on the results.
Due to computational constraints, two restrictions were imposed when generating the target signals.
First, regarding the target polynomial order $k$, the target signal was generated using only the $k$-th order term of the target polynomial without including the product of target polynomials.
Second, although the target polynomial delay time step $\tau$ should ideally be treated as the maximum delay time step, it was instead simply considered a $\tau$-step delay.
Under these constraints, the combinations of $k$ and $\tau$ were set as $\{(k_i, \tau_i)\} = \{(1,0),(1,1),(1,2),\ldots,(1,15),(2,0),(2,1),\ldots,(6,14),(6,15)\}$.
In this evaluation, the target signals were generated using Legendre polynomials as the target polynomials.
The theoretical total IPC of the reservoir is equal to the number of nodes connected to the readout layer \cite{Kubota2019}.
Since \textit{Delay} method does not directly affect the readout layer, it does not affect the total IPC. In other words, if the IPC for one component increases, the IPC for another component must decrease, indicating that the total IPC remains constant while its distribution changes. Therefore, in this evaluation experiment, we can analyze the impact of \textit{Delay} method on the nonlinearity and the memory capacity of RC by observing how the distribution of IPC changes.

\begin{figure}[t]
  \centering
  \includegraphics[width=0.95\linewidth]{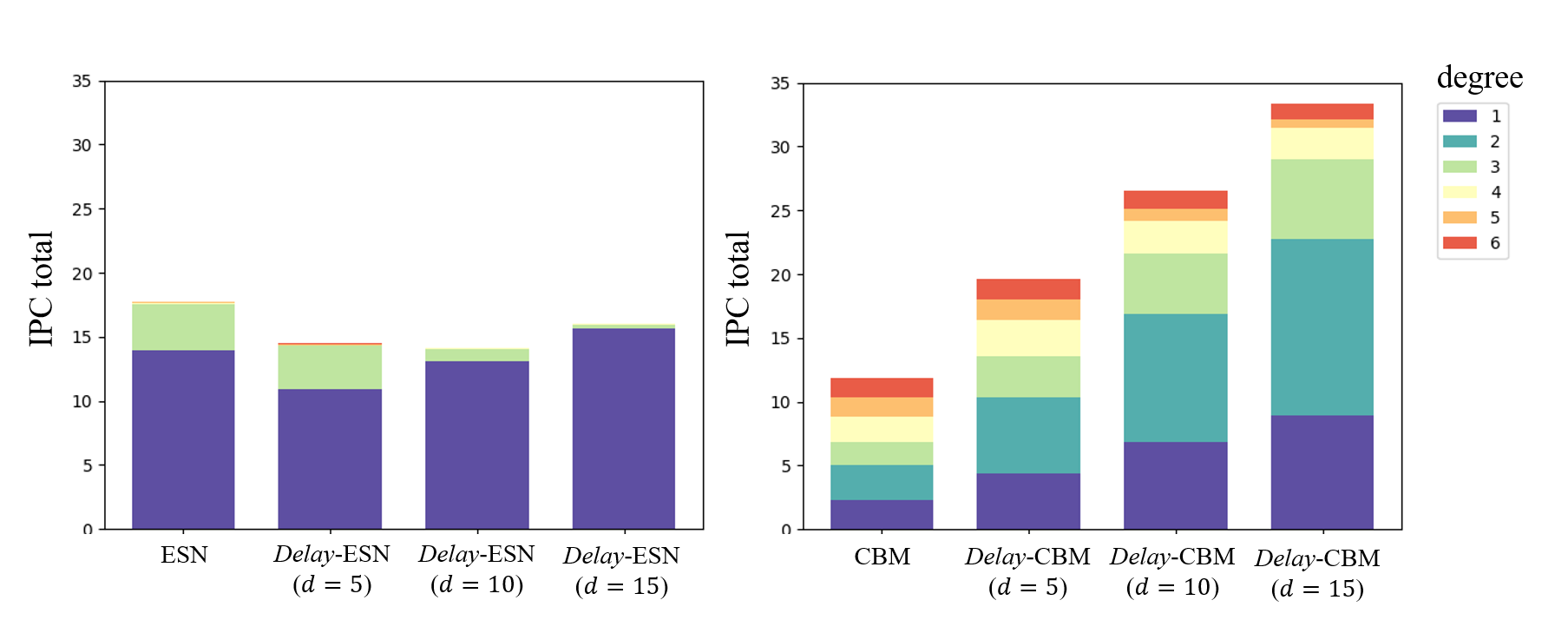}
  \caption{Evaluation results for IPC.}
  \label{fig:ipc_result}
\end{figure}

The results of the IPC evaluation are shown in Figure \ref{fig:ipc_result}. 
The model parameters used in the experiments were $(\alpha_\mathrm{in},\alpha_\mathrm{rec},\beta_\mathrm{rec})=(1.0,1.0,0.1)$ for ESN and $(\alpha_\mathrm{in},\alpha_\mathrm{rec},\beta_\mathrm{rec},\alpha_i)=(0.25,0.2,0.1,0.6)$ for CBM-RC.
For the ESN, it was observed that as the number of delay steps $d$ increased, the lower-order IPC values increased, while the higher-order IPC values decreased, resulting in a decrease in the total IPC value within the calculated combination.
This suggests that adding linear components by the \textit{Delay} method causes the higher-order IPC values to decrease significantly in the ESN.
Furthermore, the near-zero values of the even-order IPC components can be attributed to using the $\tanh$ activation function in the ESN since \rm{tanh} is an even function. 
In contrast, for CBM-RC, a highly nonlinear RC model, the total IPC increased as the number of delay steps $d$ increased, with a gradually slow growth rate for higher-order IPC components.
In both models, the increase in lower-order IPC values due to the \textit{Delay} method explains the observed performance improvement in the NARMA task, which does not require strong nonlinearity.

\begin{table}[t]
\centering
\caption{Parameters for evaluation with IPC}
\label{table:ipc_parameters}
\begin{tabular}{ll}
\hline
\textbf{Parameter} &  \\ \hline \hline
Data length $N$ & 200,1000,2500,5000,7500,10000,20000 \\
Input distribution & Uniform Distribution  \\
Target polynomial & Legendre Polynomial \\
Degree of target polynomial $k$ & 1,2,3,4,5,6 \\
Delay time steps of target polynomial $\tau$ & 0,2,...,15 \\
Number of delay steps of \textit{Delay} method $d$ & 5,10,15 \\ \hline
\end{tabular}
\end{table}

\section{Conclusion}
\label{sec:conclusion}
In this study, we propose novel methods for enhancing the memory capacity of reservoir computing (RC) solely by modifying the network configuration without changing the RC model itself.
The benchmark results using the NARMA task demonstrated that the proposed method effectively increases the memory capacity of RC. Among the methods, the \textit{Delay} method was shown to significantly contribute to this improvement.
The evaluation using IPC revealed that the \textit{Delay} method induces a change in the IPC distribution, where higher-order components decrease while lower-order components increase. Furthermore, it was shown that these changes in the IPC distribution can be adjusted by modifying the number of delay steps.
These results indicate that while there is a trade-off between memory capacity and nonlinearity in RC, the proposed methods enable adjustments to this trade-off, securing the memory capacity required for each task.
Future challenges include investigating methods to adjust the trade-off when nonlinearity is more critical than memory capacity and evaluating the effects of the \textit{Pass through} method and the \textit{Clustering} method on the IPC distribution.

\section*{Acknowledgments}
This paper is based on results obtained from a project, JPNP16007, commissioned by the New Energy and Industrial Technology Development Organization (NEDO).
This work was supported by JSPS KAKENHI Grant Numbers 23H03468, 23K18495.
This work was supported by JST ALCA-Next Grant Number JPMJAN23F3.

\bibliographystyle{unsrt}

\end{document}